\pdfoutput=1

\documentclass[11pt]{article}

\usepackage{ACL2023}

\usepackage{times}
\usepackage{latexsym}

\usepackage[T1]{fontenc}

\usepackage[utf8]{inputenc}

\usepackage{microtype}

\usepackage{inconsolata}

\usepackage{graphicx}

\usepackage{amsmath}
\usepackage{amssymb}
\usepackage{amsfonts}

\usepackage{comment}

\usepackage{booktabs}
\usepackage{multirow}
\usepackage{defs}

\title{Grounding Characters and Places in Narrative Texts}

\author{Sandeep Soni \and Amanpreet Sihra \\ University of California, Berkeley \\ \texttt{sandeepsoni,amisihra@berkeley.edu}
\And
Elizabeth F. Evans \\ Wayne State University \\ \texttt{e.f.evans@wayne.edu}
\AND
Matthew Wilkens \\ Cornell University \\ \texttt{wilkens@cornell.edu}
\And
David Bamman \\ University of California, Berkeley \\ \texttt{dbamman@berkeley.edu}
}

\begin{document}
\maketitle
\begin{abstract}
Tracking characters and locations throughout a story can help improve the understanding of its plot structure. Prior research has analyzed characters and locations from text independently without grounding characters to their locations in narrative time. Here, we address this gap by proposing a new spatial relationship categorization task. The objective of the task is to assign a spatial relationship category for every character and location co-mention within a window of text, taking into consideration linguistic context, narrative tense, and temporal scope. To this end, we annotate spatial relationships in approximately $2500$ book excerpts and train a model using contextual embeddings as features to predict these relationships. When applied to a set of books, this model allows us to test several hypotheses on mobility and domestic space, revealing that protagonists are more mobile than non-central characters and that women as characters tend to occupy more interior space than men. Overall, our work is the first step towards joint modeling and analysis of characters and places in narrative text.
\end{abstract}
\section{Introduction}
\label{sec:introduction}

The association between characters and the places they navigate is central to a wide range of literary phenomena: \emph{Bildungsromane} depict a character's journey across geographic space as a component of their psychological coming of age~\citep{Bakhtin_1987,Jeffers_2016}; the \emph{fl\^{a}neur}, who walks and observes throughout a city, epitomizes the power that follows from peripatetic access to public spaces~\citep{Benjamin_2002,Wolff_1985,Wilson_1992}; class, gender, and racial associations can render surprising or scandalous a character's mere presence in an otherwise innocuous location.
    
\begin{figure}
    \centering
    \includegraphics[scale=.6]{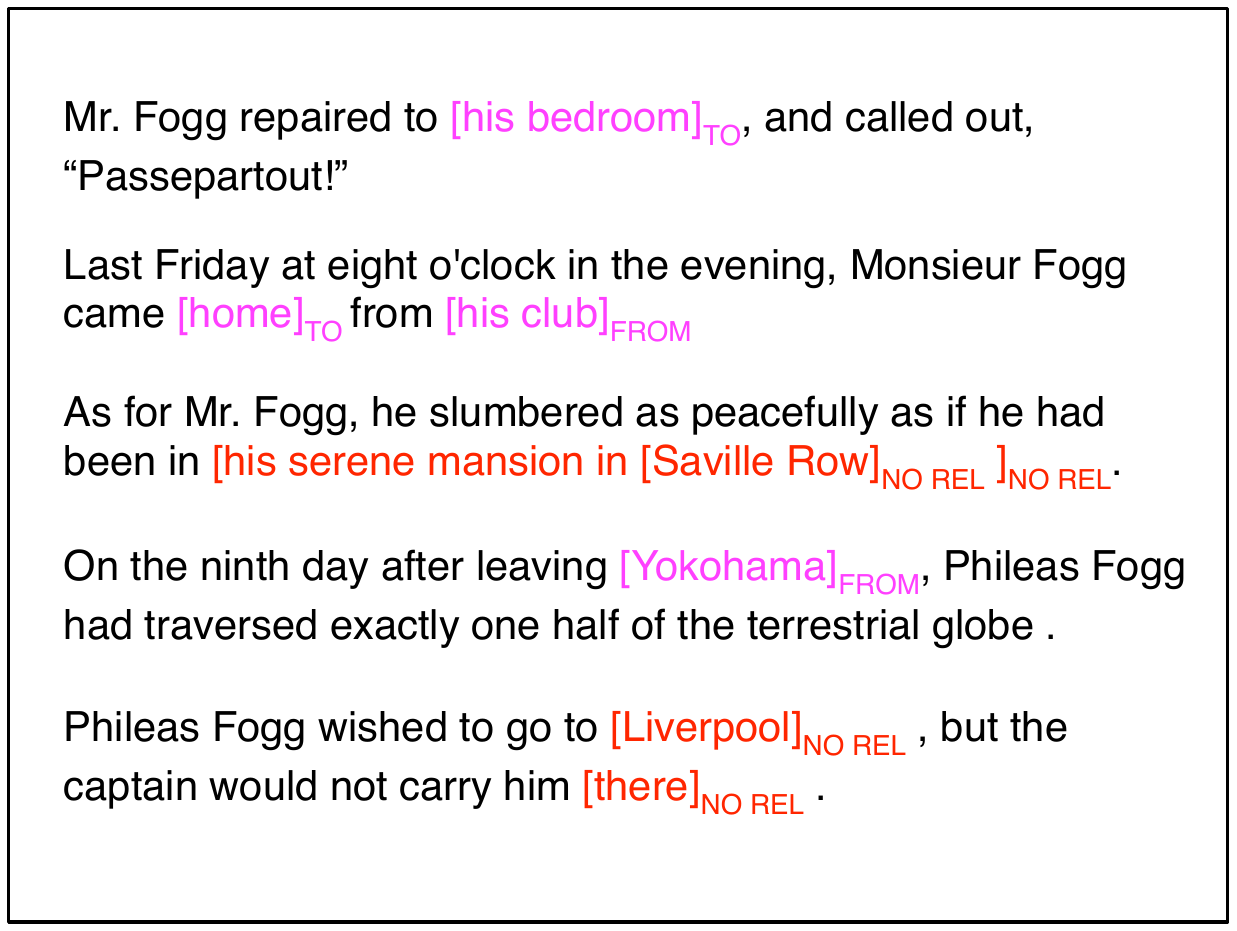}
    \caption{In \emph{Around the World in 80 Days}, Phileas Fogg is mentioned in conjunction with a wide range of places, but is only physically grounded in some. Disentangling the places he is \emph{in} from the places he is \emph{not} is crucial for tracking his movement throughout this work.}.
    \label{fig:p1-example}
\end{figure}

While much work in literary history and theory has explored this interaction, it has remained out of reach for empirical observation and large-scale comparisons with social constructs such as gender, social status~\citep{cresswell2012production} and agency~\citep{sen1993capability}.   In order to explore these questions empirically, we not only need to know the characters and places that exist within a narrative~\citep{piper-etal-2021-narrative}, but specifically how they interact: when is a character depicted as being \emph{in} a place?

\begin{figure*}
    \centering
    \includegraphics{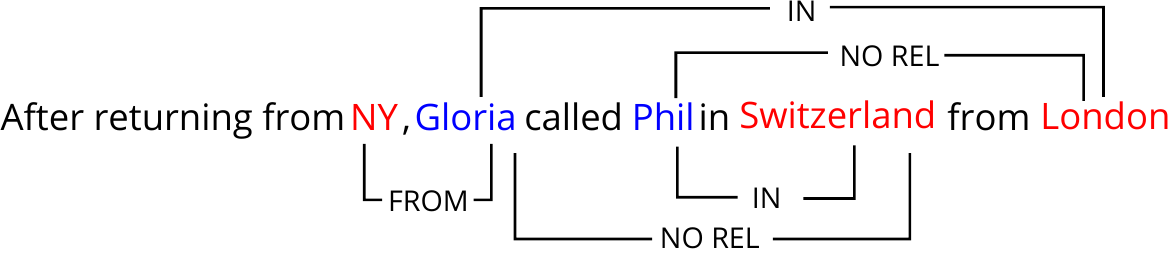}
    \caption{Example sentence to illustrate the different relationships characters and locations can hold. Characters are marked in red whereas the locations are marked in blue. The objective is to label each pair of character and location co-mention. The definition of the spatial relationships can be found in~\autoref{sec:task}}.
    \label{fig:headline-example}
\end{figure*}
NLP research has made substantial progress in the individual components of this endeavor: under entity tagging, spans of text that correspond to entities are identified and categorized by their entity types~\citep[e.g.,][]{bamman-etal-2019-annotated, hamdi2021multilingual}; methods can  ground textual spans referencing a location to their real-world coordinates~\citep[e.g.,][]{roller-etal-2012-supervised}; and extensive schemas to precisely describe the relationship between locations have been proposed~\citep[e.g.,][]{mani2010spatialml,pustejovsky2011iso}. Yet, despite this progress an important gap still remains: the technology supports the identification of characters and locations in text but falls short when it comes to placing the character with respect to a given location at any time in the story.

In this work, we address this gap by proposing a classification task whose objective is to determine the spatial relationship between a candidate character and candidate location. The classification task helps make a judgment about the nature of the spatial relationship between the character and the location, allowing us to differentiate between scenarios in which the character is, for instance, at the location, approaching a location, has left a location, or has no relationship with the location in question. To illustrate the point even further, consider the toy example in~\autoref{fig:headline-example}, in which multiple characters and locations are mentioned and their spatial relationships are annotated under the task we propose.

To make progress on this task, we annotate excerpts from books and use this annotated dataset to construct a predictive model. In applying this predictive model to a larger set of books, we test two hypotheses on mobility and domestic space, finding that protagonists are often depicted as being more mobile than other characters, and finding a strong gender effect on the kinds of spaces that are accessible: women as characters are more likely to occupy indoor or domestic spaces compared to men.

Overall, our contributions in this paper can be summarized below.

\begin{itemize}

\item We propose a new task to ground characters to locations in the story. The proposed task is an instance of a multi-class classification task with classes denoting the spatial relationship between the candidate character and location.  

\item We provide an in-depth annotation scheme for the different classes in the task and approximately 2500 annotated examples, which we openly release for public use.\footnote{The code and data for this paper can be found at \url{https://github.com/sandeepsoni/mobility-books}}

\item  We operationalize the construct of mobility and spatial positioning with the help of a trained model. We use this to test macro-level hypotheses about the mobility and centrality of the characters, in the process corroborating known claims and providing quantitative evidence for previously unverified claims.   
\end{itemize}

\section{Task}
\label{sec:task}
Given a selection of narrative text containing a mention of a character and a mention of place,
our overall task is to determine the nature of the relationship between that character and place at that instant in the story. Formally, consider a piece of text as a sequence of tokens ${w_1, w_2, \dots, w_n}$. An entity tagger identifies $C$ as the character mention spanning  tokens $w_c$ to $w_{c+k}$, where $1\leq c \leq c+k \leq n$. Similarly, the entity tagger also identifies $L$ to be the place mention spanning  $w_l$ to $w_{l+m}$, where $1\leq l \leq l+m \leq n$. Both $C$ and $L$ are within 10 tokens of each other, \ie{}, if $c+k > l$, then  $c+k-l \leq 10$; otherwise, $l+m-c \leq 10$. 

To operationalize the task further, we use entity definitions from~\citet{bamman-etal-2019-annotated}. Characters are defined as entities of the type \enttype{PER}, which include instances that refer to a single person or a group. Places or locations are defined as entities that are natural locations (\enttype{LOC}) such as \emph{the forest} or \emph{Mars}, human-built structures (\enttype{FAC}) such as \emph{the kitchen} or \emph{the church}, and geo-political entities (\enttype{GPE}) such as \emph{London} or  \emph{the village}.  Places may be entities that exist in the real world (with attendant latitude/longitude coordinates), common noun phrases that lack such geolocation, and places that exist only within imagined worlds (e.g., \emph{Hogwarts}).

\begin{table*}[]
    \centering
    \begin{tabular}{l|l|r}
    \toprule
        Category & Short description & Annotations\\
    \midrule
        IN & $C$ is at, in, or on the place $L$ & $868$\\
        NEAR & $C$ is in proximity of $L$ but not at $L$ & $184$ \\
        THRU & $C$ is passing through $L$ & $41$\\
        TO & $C$ is moving towards and is certain to reach $L$ & $171$\\
        FROM & $C$ was at $L$ before but has moved & $98$\\
        NO REL & $C$ and $L$ have no relationship & $622$\\
    \bottomrule
    \end{tabular}
    \caption{Short description for each individual category in the spatial relation identification subtask.}
    \label{tab:spatial-task-definitions}
\end{table*}


We decompose the overall task into four staged sub-tasks described next with examples. In each example, the characters marked by an entity tagger are underlined with a straight line, whereas places are marked with a wavy underline.

\subsection{Identifying groundable characters and places}
\label{sec:filtering-task}Not all people and place mentions represent entities that can be grounded with respect to each other; one important category that cannot are generic mentions~\cite{reiter-frank-2010-identifying}, such as \emph{a private establishment} below:

\begin{quote}{Reeve, \emph{The Soul Scar}}
 He insists that it must be from \place{a private establishment}.
\end{quote}

Here, \emph{a private establishment} refers to a class of entity, rather than an specific establishment that might exist in the narrative world.
As a preprocessing step, we filter out examples whose target character or location is not able to be grounded, whether through being a generic mention or through an error in entity tagging.
We formalize this as a binary classification task. An example is considered \textit{valid} if it is correctly tagged by the entity tagger and both the character and the place entity are groundable; if either of the condition fails then the example is considered \textit{invalid}. In the sample we annotated (described in ~\autoref{sec:annotation}), we found 20.8\% (522 out of 2506) examples to have the invalid label.








\subsection{Spatial relationship categorization}
\label{sec:spatial-relationship-categorization-task}
Our core task seeks to categorize the relationship between the character and the place. In total there are $6$ categories whose definitions are given in~\autoref{tab:spatial-task-definitions}, with examples in~\autoref{tab:spatial-task-examples}. We formalized these categories to be sufficiently expressive about the different scenarios after initial rounds of grounded coding on small samples taken from books.

\begin{table*}[]
    \centering
    \begin{tabular}{l | p{13cm}}
    \toprule
        Category & Example\\
    \midrule
        IN & \character{Mr. Warner} stood there, his enormous bulk seeming to fill \place{the corridor}. \\
        NEAR &But \place{the swollen waters of the river} bar our progress. \character{I} would pay its weight in gold for a raft that would transport us to the other side! \\
        THRU & If you will not dance with me again, will you walk through \place{the rooms}? ``Many admiring glances followed \character{them}--a handsomer pair was seldom seen.\\
        TO & On his return to \place{his room}, one day, \character{he} found a glass dish on the table. \\
        FROM &  Mrs Buzzby intimated her wish, pretty strongly, that the neighbours should vacate \place{the premises}; which \character{they} did, laughingly.\\
        NO REL & \character{I} know where \place{Mr. Peregrine's house} is.  \\
 \\
    \bottomrule
    \end{tabular}
    \caption{Examples for each individual category in the spatial relation identification subtask.}
    \label{tab:spatial-task-examples}
\end{table*}

\paragraph{IN.} This category is a direct judgment of whether a character is contained by the spatial boundaries delimiting the place. This relation may be evoked explicitly by prepositions such as \emph{at, in} or \emph{on}; more commonly, it must be inferred, as in the example provided in~\autoref{tab:spatial-task-examples}.

\paragraph{NEAR.} This label denotes whether a character is close to a location, but not contained within it.  This judgment is highly contextual and relative; much like discourse can compress and decompress two entities to be more or less similar for the purpose of coreference~\cite{recasens-etal-2010-typology}, so too can the discourse shorten and lengthen the apparent proximity of a character to a place.

\paragraph{THRU.} Unlike previous categories in which characters are more likely to be stationary, this category implies motion through a place. \enttype{THRU} generally implies that an \enttype{IN} relation holds as well, but provides a more specific view on the nature of that relation. Crucially, this category entails that the origin and the final destination of the character are different from the place they are marked to be passing through.  

\paragraph{TO.} This category describes a character in motion towards a destination, where we are meant to draw the inference that the destination has certainly been reached. Like \enttype{THRU}, this category generally entails an \enttype{IN} relation with the destination, but provides more specificity in the nature of that movement.

\paragraph{FROM.} This category captures movement, where a character was \enttype{IN} a place and has moved away from it.

\paragraph{NO REL.} At the core of our work is a goal of differentiating character/place pairs that co-occur in the text but that do not assert that a spatial relation holds between them.  \enttype{NO REL} describes this lack of a relation, including the cases where there is not sufficient information to deduce the relationship between $C$ and $L$. \enttype{NO REL} can apply when a character is moving towards a destination but it is uncertain if the destination has been reached or if there is epistemic narrative uncertainty expressed within the text --- where the narrator or characters do not know the relation between $C$ and $L$.




As the examples in \autoref{tab:spatial-task-examples} illustrate, the relations between a character and place are often very obliquely expressed, relying very strongly on a reader's inference rather than explicit spatial signals within the text (a point we take up again in \autoref{sec:annotation} below).

The next two subtasks consider the time for which the spatial relationship exists (\autoref{sec:temporal-span-classification-task}) and the current status of the relationship with respect to the narrative time (\autoref{sec:narrative-tense-classification-task}).

\subsection{Temporal span classification}
\label{sec:temporal-span-classification-task}
Characters may have a short-term or long-term spatial relationship with a place: when deciding whether a character is \enttype{IN} their primary home or city, for example, we can differentiate whether they are physically present there at a given moment or whether they have a habitual relationship with that place (but not necessary a punctual one at that instant).

To capture this, we mark this distinction by indicating the temporal span of the spatial relationship. The temporal span measures the amount of time that a character experiences with the place. If the relationship is short-lived or instantaneous then it is termed \textit{punctual}, as in the following example:

\begin{quote}{Altsheler, \emph{The Rulers of the Lakes}}
\character{He} was on \place{the lawn}, among the shrubbery.
\end{quote}



On the other hand, if the relationship is long-term, seasonal, or recurring, then it is considered as \textit{habitual}, as in the following example.

\begin{quote}{Sheldon, \emph{The Masked Bridal}}
Emil Correlli flew to the nearest telegraph office and dashed off a message to a \character{\place{New York} policeman}, with whom he had had some dealings while living in that city.
\end{quote}

In this example, while the text does not state whether the specific \emph{New York policeman} is \enttype{IN} \emph{New York} at the moment of utterance, we can draw the inference that they have a habitual relation to it.



\subsection{Narrative tense classification}
\label{sec:narrative-tense-classification-task}
We also want to differentiate interactions between the characters and the location as taking place in the narrative past or present. On face value, this might appear as just marking the tense of the sentence but more precisely this requires marking the tense relative to the time of narration. For example, consider the following:

\begin{quote}{Tagore, \emph{The Hungry Stones and Other Stories}}
He would dream night after night of his village home, and long to be back there. \character{He} sat in \place{the parlor} remembering the glorious meadow where he used to fly his kite all day long; \place{the broad river-banks} where \character{he} would wander about the livelong day singing and shouting for joy.
\end{quote}

In this case, the overall narration is happening in the past tense.  Within this narrative time frame, ``He sat in the parlor'' is contemporaneous with this frame and hence is considered an \textit{ongoing} relationship.  However, the narration involves a reminiscence of an event that took place at some time previous to the current narrative time (``wandering the river-banks''); the relation between \emph{he} and \emph{the broad river-banks} is hence considered an \enttype{IN} relationship that has \textit{already happened}.

\section{Annotation}
\label{sec:annotation}

We apply this framework to a sample of English-language books from Project Gutenberg. These titles span multiple centuries, index different genres, and contain a mix of fiction and narrative non-fiction.  All works are in the public domain in the United States and are able to be openly published along with our annotations.

From this collection, we apply the LitBank entity tagger to identify all person and place mentions, and sample passages containing at least one character and one location separated by 10 or fewer tokens. 2506 of these samples were annotated in total by 3 annotators. The annotation process started by first carrying out pilot annotations. 
After an initial round of annotations, a codebook was created which was further refined in each subsequent pilot annotation round. Next, the codebook was formalized into an annotation guideline document. The document described the tasks, defined the categories per task, and gave intuitive and real examples for each category. The annotation guidelines were iteratively refined throughout the annotation process. 

\paragraph{Training.} Every annotator, other than the lead author, underwent training by reading the annotation guidelines and getting  familiarized with the task. A small random sample of 50 examples were annotated and the annotations were discussed with the lead author. At this stage, any disagreements were discussed, discrepancies in the guideline were corrected, and additional clarification, if any, was added to the guideline. During training, the annotators were specifically asked to manage each annotation in under 2 minutes. 

\paragraph{Interannotator agreement.}After training, a common randomly picked sample of 261 examples was annotated independently by 2 annotators, yielding a Cohen's $\kappa$ of 0.53 on identifying the validity of entities, 0.58 on the spatial relationship categorization task, 0.48 on the temporal span classification task, and 0.53 on the narrative tense classification task. After this phase, every annotator separately carried out the annotations.  During this separate annotation, difficult and ambiguous examples were marked and discussed by all three annotators. The annotation guidelines were refined if necessary. Any remaining disagreements  were resolved by the lead author. A distribution of the labels in the annotated data is given in~\autoref{tab:spatial-task-definitions}.

\subsection{Annotation challenges}

As the examples in \autoref{tab:spatial-task-examples} make clear, along with our moderate agreement rate,  spatial relationship classification  is a challenging task that extensively draws on inference rather than overt lexical cues. The SpatialML task, which in many respects is conceptually similar to our task, also has low inter annotator agreement, pointing, in general, to the difficulty in relating locations with other entities~\citep{mani-etal-2008-spatialml}.

To make a judgment on an example, an annotator has to make several inferences. They have to draw upon world knowledge to avoid false positives in recognizing characters and locations; for instance, the flower ``Lily of the Valley'' should not be mistaken for a reference to a character (or containing a reference to a specific place). In some cases, an annotator has to perform common sense reasoning; for instance, ``looking out of the kitchen window’’ has a common sense implication of being inside the kitchen. Furthermore, narrative texts are replete with dialogues that implicitly need to be disentangled in order to attribute locations to characters. Similarly, to correctly link characters to locations, entity coreference---naturally found in narrative texts---needs to be tracked carefully.








\section{Model}
\label{sec:model}
We build classifiers for different tasks using the annotated data. Every annotated example consists of a span of tokens denoting a character and a span of tokens denoting a location. We use the token representations from a BERT language model~\citep{devlin-etal-2019-bert}, which we then aggregate into span representations by averaging the token representations. Assuming that $\cbf$ and $\lbf$ are vector representations for the character and location respectively, the classification model is simply:

\begin{equation*}
    \text{Softmax}(\text{Feedforward}(\cbf \oplus \lbf)), 
\end{equation*}

where $\oplus$ is the concatenation operator between vectors. The feedforward network stacks linear layers with sigmoid activation between the layers.

\paragraph{Setup.}
We train a binary classification model to identify groundable character/place pairs, temporal span classification and narrative tense classification, and a multi-class classification model for spatial relationship classification. 
To assess accuracy, we divide all the annotations into 70\% for training, 10\% for tuning the hyperparameters, and the remaining 20\% for testing. Features of the classification model are concatenated span representations from the final hidden layer of the \texttt{bert-base-cased} language model~\citep{wolf-etal-2020-transformers}. 
We train our models to minimize the cross entropy loss, and maintain a learning rate of $1e^{-5}$.
We find the optimum value of the hyperparameters by testing the accuracy of the model against the development set. Specifically, we tune the number of epochs to train the model for early stopping from the set $\{1 \dots 15\}$, the number of hidden layers from the set $\{0,1\}$, and length of each excerpt passed to the model, operationalized as the number of tokens from the set $\{10, 50, 100\}$ before the first and last mention of the character and place in the excerpt. 


\paragraph{Results.}

We compare the performance of a BERT-based classifier with several alternatives.  First, we create a simple \textbf{majority class} baseline by assigning the most frequently occurring label in the training set. This baseline assigns the same label to each example.  We also compare the performance with two large language models, \textbf{ChatGPT} and \textbf{GPT-4}\ \citep{gpt4}, in order to explore the degree to which prompting alone---as distinct from optimizing directly on the task---is able to recover information about this complex phenomenon; such LLMs may offer promise for research in cultural analytics and computational social science by reducing the necessity for large amounts of training data~\citep{underwoodtime2023,ziems2023css}, though recent work has urged caution in using these models with literary texts~\citep{chang2023speak}. 
For each of the two, we created prompts that included the description of the task, a few examples, and a rationale for each of the labels. 

\begin{table}[]
    \centering
    \begin{tabular}{p{1.4cm}|r|r|r|r}
    \toprule
    Model &Valid&Rel.&Temp.& Tense\\
    \midrule
    Majority class & 79.8 & 43.5 & 56.5 & 71.0\\
    \midrule
    ChatGPT & 47.7 & 30.2 & 31.4 & 70.7 \\
    \midrule
    GPT-4 & 83.8 & 51.0 & 59.3 & 69.2 \\
    \midrule
    BERT classifier & 87.4 & 56.8 & 73.3 & 79.0\\
    
    \bottomrule 
    \end{tabular}
    \caption{Accuracy of the model on each task (whether a character/place pair is valid; classifying spatial relationship, temporal span and narrative tense) shown in percentage; The majority class model for every task predicts the most common  label in the training set. The performance is reported on the test set.}
    \label{tab:modeling-results}
\end{table}

\begin{figure}
    \centering
    \includegraphics[width=\linewidth,scale=1.2]{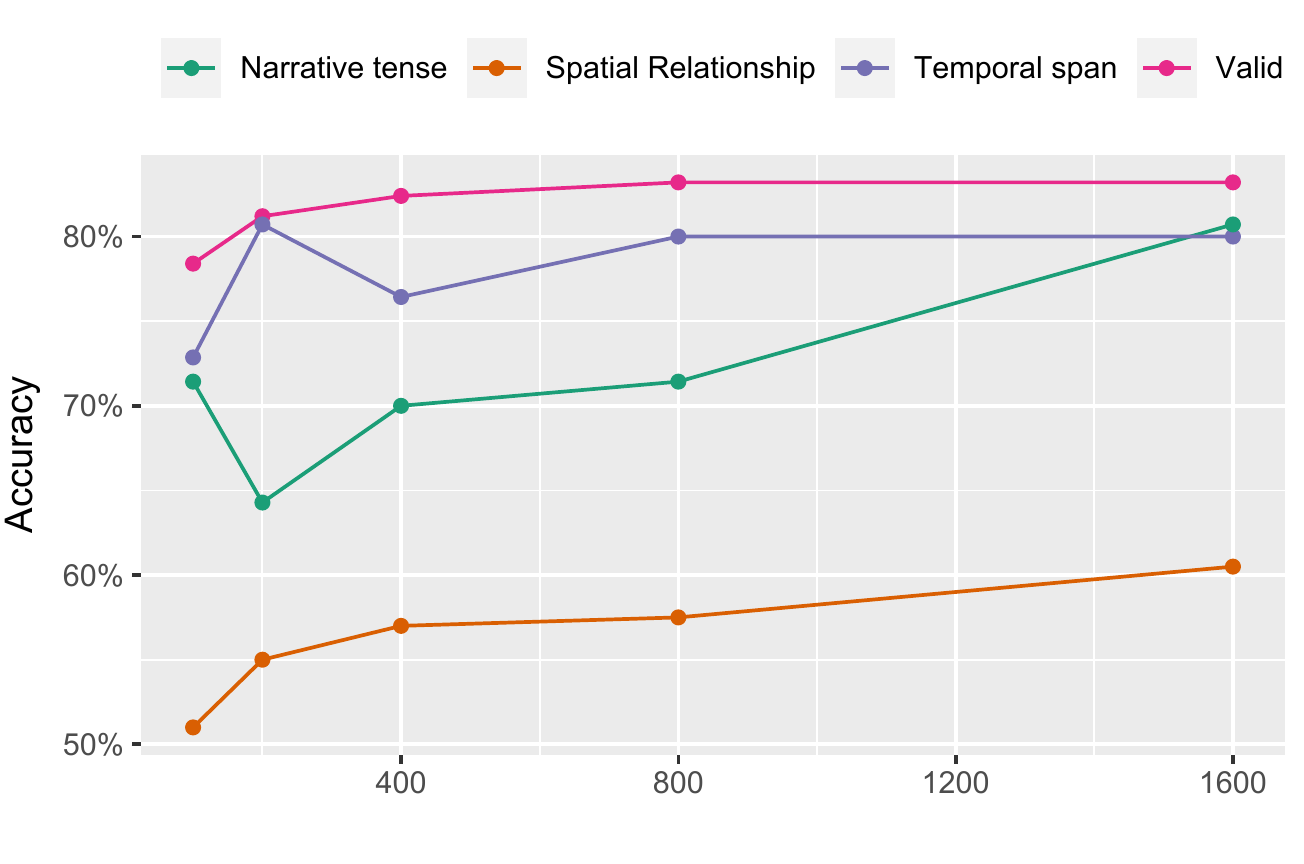}
    \caption{BERT classifier's performance as a function of the size of the training set.}
    \label{fig:training-size}
\end{figure}

The overall results are presented in~\autoref{tab:modeling-results}, which illustrate the comparative advantages of optimizing directly on the task. BERT improves over the simple majority classifier on all tasks, and also demonstrates an
improvement over GPT-4 by close to 6 points on the spatial relation classification. This performance gap can be attributed to the fact that ChatGPT and GPT-4 is a limited in-context learning setup whereas BERT is trained on many examples. To elaborate this further, we show the accuracy of the BERT classifier as a function of the size of the training set in ~\autoref{fig:training-size}. We find that expanding the training set with more examples benefits the classifier on all the tasks. 

Finally, we break down the performance statistics by each category for this task in~\autoref{tab:spatial-modeling-results}. As expected, the model struggles to make predictions about rare classes suggesting the need to annotate more data. The classifier's F1 is highest for the two most prominent classes (\enttype{IN} and \enttype{NO REL}) and we rely on these for the analysis that follows.     

\begin{table}[]
    \centering
    \begin{tabular}{l|r|r|r}
    \toprule
         Category & Precision & Recall & F1\\
    \midrule
         IN & 0.68 & 0.61 & 0.65 \\
         NO REL & 0.54 & 0.68 & 0.60 \\
         TO & 0.40 & 0.53 & 0.45 \\
         FROM & 0.43 & 0.45 & 0.44 \\
         NEAR & 0.50 & 0.24 & 0.32\\
         THRU & 0.20 & 0.33 & 0.25 \\
    \bottomrule
    \end{tabular}
    \caption{The performance of the BERT-based classifier on the spatial prediction task. The table shows the performance broken down per class.}
    \label{tab:spatial-modeling-results}
\end{table}

\section{Analysis}
\label{sec:analysis}
A predictive model of the spatial relationships can be used to draw inferences for individual character and location co-mentions but it also opens up the possibility of macro-analysis by aggregating these individual predictions across books. In this section, we show how the model can be applied to test known hypotheses and discover new findings about the mobility of characters in books.

\paragraph{Dataset.} To measure the association between characters and place, we draw on several textual sources, each approximately 100 novels: the collection of English-language books from Project Gutenberg that form  LitBank~\citep{bamman-etal-2019-annotated}, spanning 1719--1922; Pulitzer prize nominees from 1923--2020 (one per year); bestsellers from the \emph{NY Times} and \emph{Publishers Weekly} from 1923--2020 (one per year); novels written by Black authors, either from the Black Book Interactive Project\footnote{\url{http://bbip.ku.edu/novel-collections}} or Black Caucus American Library Association award winners from 1928--2018; works of Global Anglophone fiction (outside the U.S. and U.K.) from 1935--2020; and genre fiction, containing science fiction/fantasy, horror, mystery/crime, romance and action/spy novels from 1928--2017.




\subsection{Protagonist Mobility}
A long-held understanding in narrative studies is that stories of development, epitomized by the \emph{Bildungsroman}, nearly always involve movement through both time and space~\citep{bachtin1981forms}. The main characters who provide the centralized focus of such narratives are generally more mobile in comparison to other characters. We quantitatively test this hypothesis by applying our model to ground characters in places, then using the model's predictions to measure the mobility of central characters in a story, which we then compare to the mobility of non-central characters.

We operationalize the distinction between protagonists and non-protagonists by the frequency of their mentions in text (selecting the single most frequent character as the protagonist and all others as non-protagonists) and the mobility of a character as the number of \emph{distinct} locations among a fixed set of location mentions at which the character is grounded (\ie{}, has prediction from the model as \enttype{IN}); this is analogous to a type-token ratio over grounded place.  We calculate mobility over the same number of fixed location mentions for all characters to remove frequency effects from our estimation of mobility.

To test our hypothesis, we pair the most central character in a book with a randomly picked character from the next $5$ central characters. We then sample $50$ \enttype{IN} predictions for each of the characters in the pair and calculate the mobility from this sample. 
Averaging the mobility across all books, we can compare the mobility of protagonists to that of non-protagonists. To control for randomness due to sampling, we repeat the process $100$ times. 

We find that the protagonist is approximately $22\%$ ($\pm 10\%$)  more mobile, on average, than the next $5$ non-protagonists. Thus, we provide positive empirical evidence for the claim that lead characters are, in general, more mobile.

We also test the hypothesis with respect to referential gender obtained using gender inference in BookNLP,\footnote{\url{https://github.com/booknlp/booknlp}} separating the books where protagonists align with \{\emph{she, her}\} pronouns from the books where the protagonists align with \{\emph{he, him, his}\} pronouns. By repeating the same procedure on this stratified set, we find slight but statistically insignificant variation in mobility across gender:  lead characters who are women are $28\%$ ($\pm 13\%$) more mobile compared to non-central characters in those books; in contrast, lead characters who are men are $19\%$ ($\pm 12\%$) more mobile than their non-central characters. This gender parity suggests that mobility is intricately linked to the leading role of the characters, independent of their gender.

\subsection{Interior space and gender}
\label{sec:interior-space-and-gender}
Are there gender differences in characters' position in space?  Prior work in literary studies has pointed to the alignment between feminized characters and domestic spaces, especially (though not exclusively) in novels that predate the Second World War~\citep{armstrong1987}.  When we are able to ground characters in the specific places they occupy, do we see this association empirically?


To test this hypothesis, we mark a total of $500$ most frequently occurring locations as either ``indoor'' (e.g., \word{his chamber}) or ``outdoor'' (e.g. \textit{the coast}). Next, we query the model's assignment of the spatial category to each character and location co-mention and filter out every spatial category except \enttype{IN}. We use BookNLP's gender inference to obtain the referential gender of each character, focusing on characters aligned with \{\emph{he, him, his}\} and \{\emph{she, her}\} pronouns.

We calculate the proclivity towards occupying indoor spaces by  gender as $P(L=\text{``indoor''} | g(C))$, where $g(C)$ gives the referential gender of the character; $g(C) = \{he, she\}$.
We find that among this set of indoor/outdoor places, women appear indoors 64\% of the time, while men appear indoors only 54\% of the time, a relative disparity of 18.5\% (\autoref{tab:indoor-proclivity}).



\begin{table}[]
    \centering
    \begin{tabular}{c|c}
       \toprule
       gender  &  indoor probability \\
       \midrule
       he/him/his  & $0.54 \pm 0.002$\\
       she/her & $0.64 \pm 0.002$\\
       \bottomrule
    \end{tabular}
    \caption{The proclivity of characters based on their referential gender to occupy indoor or domestic spaces. Both men and women as characters tend to occupy indoor spaces, possibly suggesting that the \emph{de facto} settings in a novel are indoor spaces; women  tend to be more indoors than men. The $95\%$ confidence intervals are calculated using a Wald test.}
    \label{tab:indoor-proclivity}
\end{table}

\subsection{Interior space and time}
\begin{figure}
    \centering
    \includegraphics[width=\linewidth]{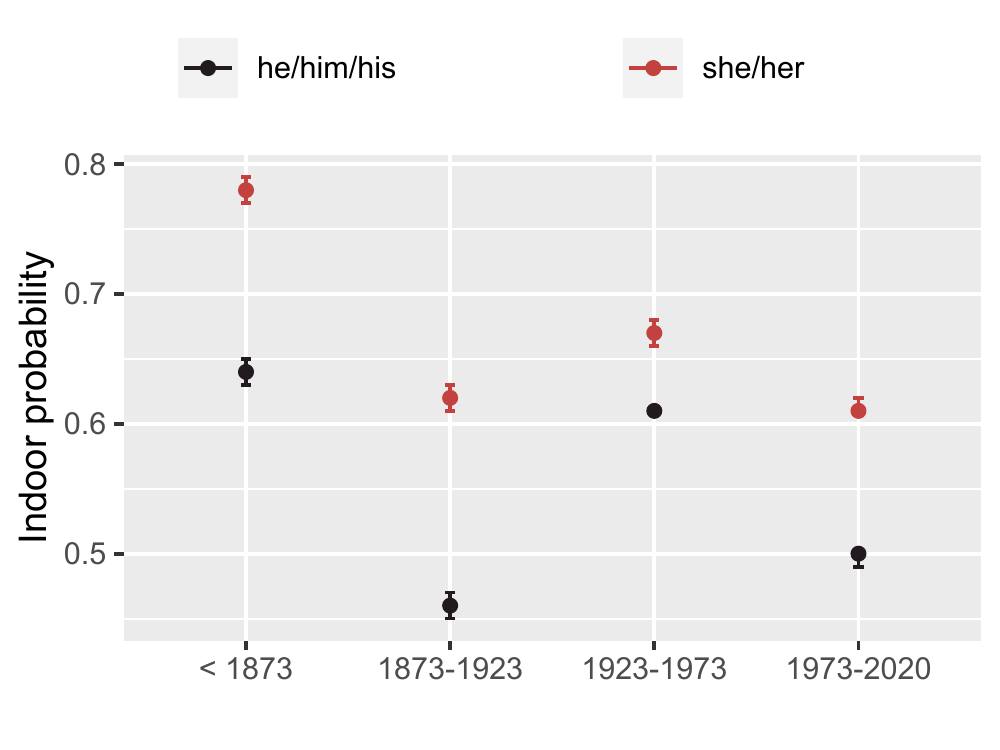}
    \caption{Proclivity of characters towards occupying indoor spaces over time. We find that characters are generally less inclined to reside in indoor spaces over time. The proclivity for both the genders to stay indoors is slowly converging towards 0.5.}
    \label{fig:indoor-proclivity-time}
\end{figure}

After establishing variation in spaces occupied by characters based on the gender, we also test if this variation exists over time. To do this, we repeat the analysis in~\autoref{sec:interior-space-and-gender}, but on temporal slices of the data: we place books into four temporal buckets (<1873; 1873-1923; 1923-1973; 1973-2020)
and calculate the association with indoor spaces for books in each temporal slice. 
The results are shown in~\autoref{fig:indoor-proclivity-time}.

We see that the tendency to be depicted indoors for characters of both genders drops over time since the end of nineteenth century; in fact, characters in books from the twentieth century who are aligned with the \{\emph{he, him, his}\} pronouns are equally inclined to occupy exterior spaces. 
The proclivity of the characters to stay indoors also shows sign of converging over time towards 0.5. An exception to this trend is the period between 1923 to 1973 which saw a surprising rise in the proclivity of characters aligned with \{\emph{he, him, his}\} pronouns to stay indoors. We leave the deeper investigation of this surprising empirical fact to future work. 

\section{Related Work}
\label{sec:related}
Our two primary contributions are a new task on classifying the spatial relationship between characters and locations and subsequent analysis about the movement of characters in narrative texts. We briefly describe the relevant work along these aspects.

In the analysis of characters and locations, named entity recognition (NER) has attracted considerable attention for its use in narrative texts such as books~\citep[e.g.,][]{brooke-etal-2016-bootstrapped,bamman-etal-2019-annotated}; a more focused task is the identification of characters from text~\citep[e.g.,][]{he-etal-2013-identification,bamman-etal-2014-bayesian,vala-etal-2015-mr}. Progress has also been made at coreference resolution, crucial for correct identification of character references in text~\citep{bamman-etal-2020-annotated}. In this work, we use a named entity tagger to mark spans of text as characters and locations, but extend the technology to categorize the relationship between a character and a location.

An analytical lens that previous research has used is to study character networks in novels~\citep[e.g.,][]{elson-etal-2010-extracting, agarwal-etal-2013-automatic,dekker2019evaluating} and films~\citep[e.g.,][]{krishnan-eisenstein-2015-youre}. Similarly, recent research has developed models of inter-character relationships in literary text~\citep[e.g.,][]{iyyer-etal-2016-feuding,srivastava2016inferring,chaturvedi2017unsupervised}. Unlike prior research, our work does not focus only on analyzing characters but instead seeks to understand their spatial relationship with places.

With respect to locations, there has been some work on toponym resolution in historial texts --- a task that computationally links the text to geographic coordinatess~\citep[e.g.,][]{smith2001disambiguating,speriosu-baldridge-2013-text, delozier-etal-2016-creating}. Others have proposed rich annotation schemes to relate multiple placenames~\citep[e.g.,][]{mani-etal-2008-spatialml,pustejovsky2011iso}. Our proposed task and its associated annotations differs from the previous work because we relate locations to characters. 

Finally, our analytical work exemplifies application of computational methods to historical, literary text. Previous work has analyzed single attributes of a character such as gender in English fiction books~\citep{underwood2018transformation}. In contrast, our analysis considers multiple attributes such as gender and spatial location. Prior work has also analyzed the dynamics of spatial locations, including mobility~\citep[e.g.,][]{evans2018nation}, and related it to measurements of emotions~\citep[e.g.,][]{heuser2016emotions,semyan2022map} and race~\citep{burgers2020familial}. The unit of such analyses have been books, in contrast to our work where we zoom in to analyze the mobility of characters within books.
\section{Conclusion}
\label{sec:conclusion}
We propose a new, challenging task of grounding characters to places in narrative texts.  Unlike other domains that rely more heavily on surface lexical features to denote spatial relationships between entities, narrative texts often rely on indirect inference, exploiting a reader's  commonsense knowledge and mental models of the geography within the imagined world to establish relationships between characters and the places they inhabit and navigate. This complexity presents challenges for both annotation and modeling, but we find that predictive models are able to ground characters in places for relations that are well attested in our data (e.g., \enttype{IN} and \enttype{NO REL} in particular), which offers promise for increasing the size of training data for less represented categories.

In focusing on the core category measuring when a character is \enttype{IN} a place, we illustrate the affordances of this work: grounding characters in place allows us not only to measure the formal elements of narrative (the structural mobility of protagonists) but also capture the degree to which mobility and the experience of space in narrative is intimately bound with gender.

\section{Limitations}
\label{sec:limitations}
Our conceptualization of the core task has some important limitations. We highlight three main limitations here. First, in order to tie a character to a place, we require that both the character and the place are explicitly mentioned in the text. This simplyfying approach helps annotation and modeling but is inadequate against the general setting of grounding any character at any time in the story. 

Another limitation with our current approach is the assumption that the location of a character is independent at every instance in the story. It is because of this assumption that we can label every character and location co-mention without considering any other labels. In reality, however, location of a character at some time is highly dependent on the location of the character at a previous time.

Finally, the spatial relationship categories are designed to be coarse. This is helpful in setting up the task as a classification task but collapses information that can be useful. For example, if a character is described to be \emph{standing outside the southern gate of a building}, our current approach will assign the \enttype{NEAR} label retaining only the aspect of distance and not the spatial orientation.
\section{Ethics Statement}
\label{sec:ethics}

While our analysis covers a wide range of English-language novels (including global Anglophone fiction, bestsellers, Pulitzer nominees, works by Black authors, genre fiction and largely canonical texts written before the 20th century), our annotated data is drawn exclusively from works in the public domain on Project Gutenberg.
Our choice of Project Gutenberg as our sole source of annotated data carries a potential risk of bias in our modeling and analysis. This is because Project Gutenberg consists of data imbalances by favoring books written in English and  predominantly by authors from the U.S. and the U.K. The exclusion of authors from other demographics continues the long-standing issue of \emph{underexposure} because of which our tools and analyses are rooted in the same language and cater to a small, highly privileged demographic~\citep{hovy-spruit-2016-social}.   

\section*{Acknowledgements}

The research reported in this article was supported by funding from the National Science Foundation (IIS-1942591) and the National Endowment for the Humanities (HAA-271654-20). We also thank Anna Ho and Mackenzie Cramer for their contributions to annotation and for various discussions that led to clarifying the operationalization of the labels.

\bibliography{anthology,custom}
\bibliographystyle{acl_natbib}


\end{document}